\documentclass{article}

\PassOptionsToPackage{numbers, compress}{natbib}

\usepackage[final]{neurips_2024}

\usepackage{graphicx}
\graphicspath{{figures/}}

\usepackage{subfiles}

\usepackage[utf8]{inputenc} 
\usepackage[T1]{fontenc}    
\usepackage{hyperref}       
\usepackage{url}            
\usepackage{booktabs}       
\usepackage{amsfonts}       
\usepackage{nicefrac}       
\usepackage{microtype}      
\usepackage{xcolor}         

\usepackage{graphicx}
\usepackage{amsmath}
\usepackage{tabularx}
\usepackage{multirow} 
\usepackage{subcaption} 
\usepackage{graphicx} 

\title{Gaze-Assisted Medical Image Segmentation}

\author{%
Leila Khaertdinova$^{1}$ \quad Ilya Pershin$^{1, 2}$ \quad Tatiana Shmykova$^{1, 2}$ \quad Bulat Ibragimov$^3$ \\
$^1$Research Center for AI, Innopolis University \\ $^2$Kazan Federal University \quad $^3$University of Copenhagen\\
\texttt{l.khaertdinova@innopolis.university}\\
\texttt{\{i.pershin,t.shmykova\}@innopolis.ru}\\
\texttt{bulat@di.ku.dk}
}

\begin{document}

\maketitle

\begin{abstract}
  The annotation of patient organs is a crucial part of various diagnostic and treatment procedures, such as radiotherapy planning. Manual annotation is extremely time-consuming, while its automation using modern image analysis techniques has not yet reached levels sufficient for clinical adoption. This paper investigates the idea of semi-supervised medical image segmentation using human gaze as interactive input for segmentation correction. In particular, we fine-tuned the Segment Anything Model in Medical Images (MedSAM), a public solution that uses various prompt types as additional input for semi-automated segmentation correction. We used human gaze data from reading abdominal images as a prompt for fine-tuning MedSAM. The model was validated on a public WORD database, which consists of 120 CT scans of 16 abdominal organs. The results of the gaze-assisted MedSAM were shown to be superior to the results of the state-of-the-art segmentation models. In particular, the average Dice coefficient for 16 abdominal organs was 85.8\%, 86.7\%, 81.7\%, and 90.5\% for nnUNetV2, ResUNet, original MedSAM, and our gaze-assisted MedSAM model, respectively. \footnote{\href{https://github.com/leiluk1/gaze-based-segmentation}{https://github.com/leiluk1/gaze-based-segmentation}}
\end{abstract}

\section{Introduction}

The workload of radiologists has increased significantly over the past several decades, leading to concerns about the impact of fatigue on the accuracy of medical image interpretation \citep{DanLantsman2022, Bruls2020, TaylorPhillips2019, Pershin2023}. To mitigate the risk of decreased diagnostic quality, several strategies have been proposed, including limiting radiologists' workloads and adjusting the pace of image analysis \citep{Alexander2022}. Another viable approach involves optimizing radiologists' workflows, encompassing everything from the layout of reading rooms to team management practices \citep{McGrath2022}. Among these strategies, Artificial Intelligence (AI) emerges as a particularly promising solution.

Along with the potential prospects of AI, there are also significant drawbacks related to the explainability of AI models \citep{Reyes2020, Borys2023} and the radiologists' bias towards AI \citep{Gaube2021, Chen2022, Li2023}. These issues introduce challenges to the automation of the radiology workflow and the seamless integration of AI. Nonetheless, the emerging paradigm of hybrid intelligence, which combines human expertise with AI capabilities, provides a viable solution. Rather than relying on AI autonomously, hybrid intelligence emphasizes collaborative interactions between radiologists and AI to optimize their workflows. One promising method within this paradigm is the use of gaze analysis, a natural form of interaction given that medical visualization is central to radiology. Recently, multiple works in the research literature investigated the use of gaze data to improve deep learning models in medical image analysis \citep{Karargyris2021, Wang2022} and to extract patterns related to the increased workload \citep{Pershin2022}.

One of the tasks frequently addressed in surgical and radiotherapy planning is medical image segmentation \citep{Li2022}. Precise segmentation masks are crucial for providing essential information that aids in clinical diagnosis and patient monitoring. Furthermore, organ segmentation is vital in therapeutic interventions, particularly in cancer treatments that rely on radiation therapy \citep{sharma2010automated, chen2021deep}. The advent of deep learning has enabled the automatic segmentation of medical images, significantly simplifying the annotation process. These methods alleviate the challenges posed by the labour-intensive nature of manual segmentation, which often demands several hours of effort from annotators for each case \citep{luo2021word}. However, segmentation inaccuracies often necessitate accurate manual corrections. Recent advances in interactive segmentation have made this correction process more efficient across various domains, including medicine, reducing the effort to just a few mouse clicks \citep{medsam_review, MedSAM, zhang2023segment}. 

\begin{figure*}
  \centering
  
  \begin{subfigure}{0.3\linewidth}
    \includegraphics[width=\linewidth]{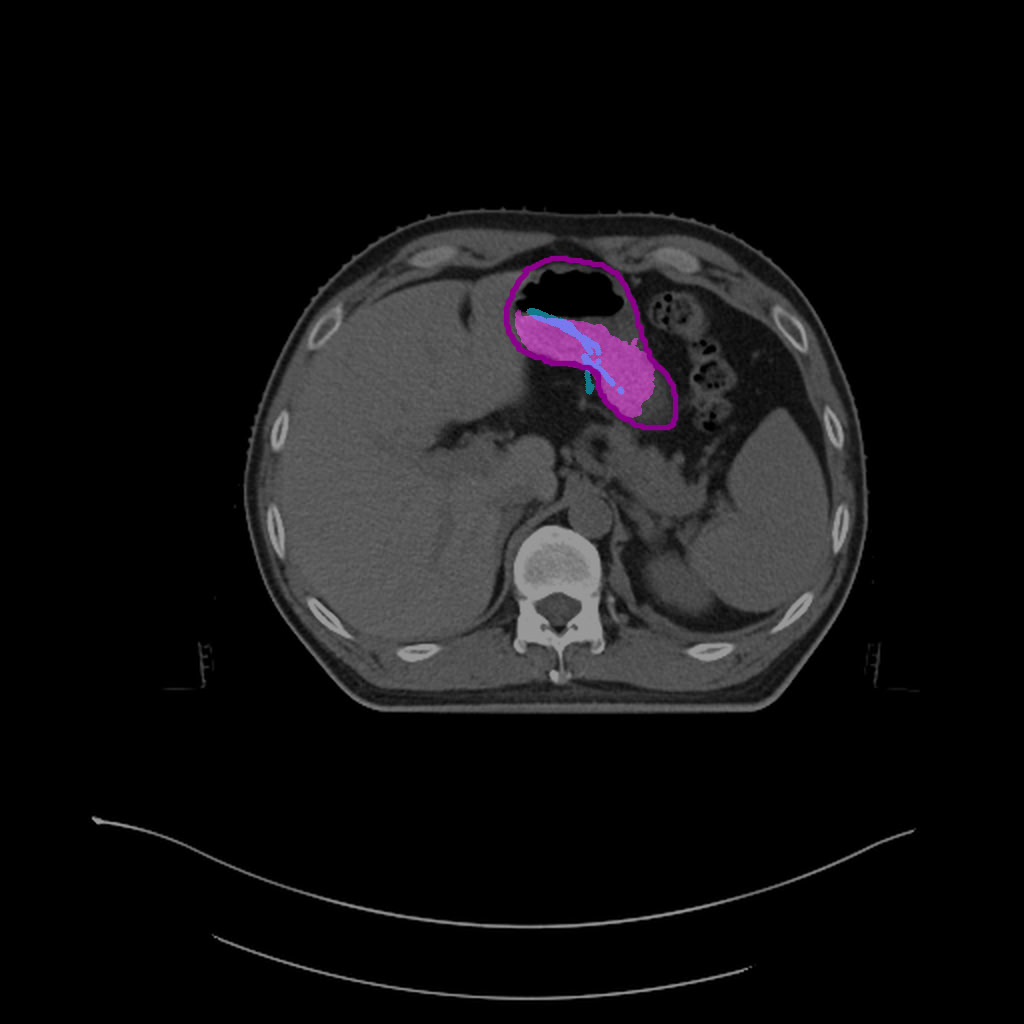}
    \caption{The initial predicted mask by gaze data.}
    \label{fig:short-a}
  \end{subfigure}
  \hfill
  \begin{subfigure}{0.3\linewidth}
    \includegraphics[width=\linewidth]{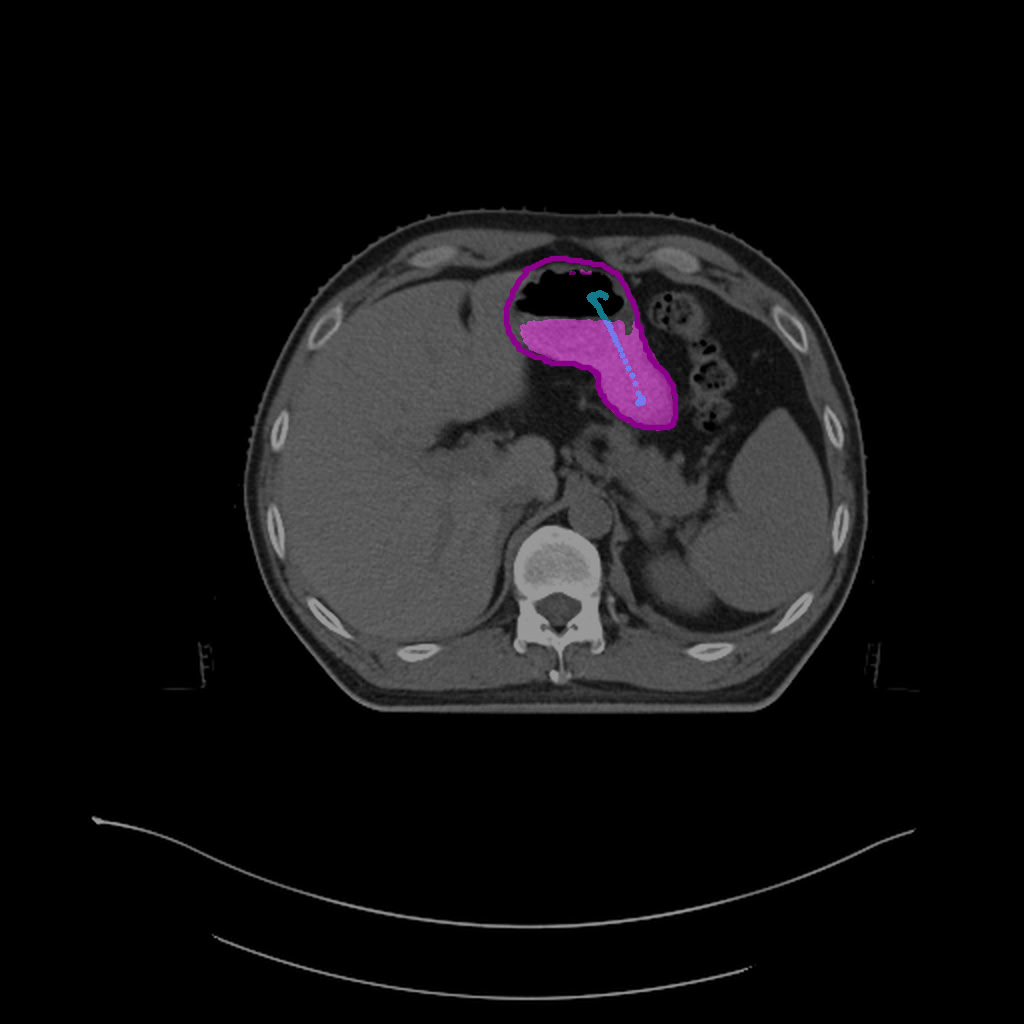}
    \caption{The mask after half the correction time by gaze data.}
    \label{fig:short-b}
  \end{subfigure}
  \hfill
  \begin{subfigure}{0.3\linewidth}
  \includegraphics[width=\linewidth]{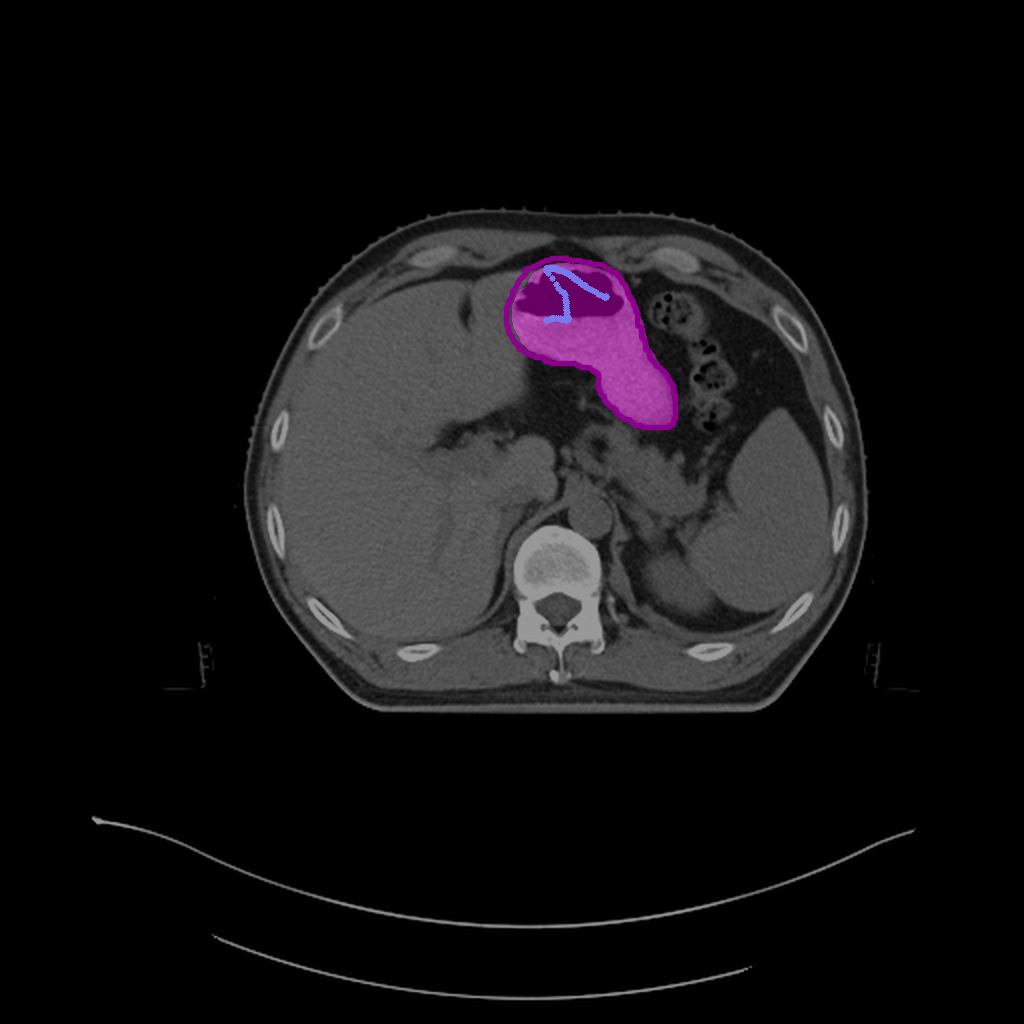}
  \caption{The mask at the end of the correction time by gaze data.}
  \label{fig:short-c}
  \end{subfigure}
  
  \caption{Steps to correct a stomach segmentation mask on a CT slice. Each subfigure shows the outline of reference segmentation contours, the predicted segmentation mask, and gaze points (blue) that are used for prediction.}
  \label{fig:vis}
\end{figure*}

In this paper, we aim to incorporate eye gaze data into the interactive segmentation workflow. Specifically, we propose lightweight fine-tuning strategies of the existing interactive segmentation model, namely Segment Anything Model, or SAM \citep{sam2023, MedSAM}, using simulated eye gaze data. At inference time, utilizing streams of real gaze data instead of mouse clicks could provide a more intuitive and faster way for annotating 3D medical images, such as CT (Computed Tomography) and Magnetic Resonance Imaging (MRI) scans, which comprise a large number of slices \citep{sadeghi2009hands}. Figure~\ref{fig:vis} illustrates an example of the gaze-based process for correcting a stomach segmentation mask on a single CT slice. 

Overall, the main contributions of this study are listed as follows:
\begin{itemize} 
    \item We present a novel approach for the real-time correction of segmentation in CT scans based on sequential gaze information. More specifically, we suggest adapting existing interactive segmentation frameworks to incorporate gaze data as a prompt.
    \item We introduce and evaluate various fine-tuning options for the medical interactive segmentation model, namely MedSAM \citep{MedSAM}, by utilizing synthetic prompts that mimic eye gaze streams.
    \item Extensive experiments and ablations conducted by both human participants and medical experts demonstrate the effectiveness of the proposed method, compared to the existing methods, such as mouse clicks and bounding boxes. Additionally, we analyze the radiologist's behaviour during the segmentation process, focusing on eye movements and the time spent on experiments using eye-tracking technology.
\end{itemize}

\section{Related work}

\textbf{Interactive segmentation.} Interactive segmentation is a specific type of segmentation that leverages user interaction to enhance the segmentation process. Usually, users provide information for initial segmentation, such as bounding boxes or textual descriptions of the region of interest, and, if necessary, subsequently refine the quality of the predicted segmentation. With its history in the pre-deep learning era, interactive segmentation has evolved from graph represented \cite{Boykov} and random walk \cite{Grady2006} based approaches to contemporary foundation models \cite{sam2023, seem}.

SAM \cite{sam2023} is the state-of-the-art interactive segmentation model, which provides impressive performance across diverse segmentation tasks that involve human interaction. The architecture of the model consists of a Vision Transformer-based image encoder, a prompt (mouse clicks coordinates, bounding boxes, textual descriptions) encoder, and a mask decoder. While SAM demonstrated impressive performance on natural images, the performance in clinical settings is weaker \cite{orig_sam_for_medical}. To address this, MedSAM \cite{MedSAM} was fine-tuned on the large-scale medical dataset and adapted to medical image segmentation using bounding boxes.

\textbf{Eye tracking technology in the medical field.} Eye-tracking technology has emerged as an effective tool in the medical field, providing diverse applications across various disciplines \citep{tahri2023eye}, particularly in image segmentation. Sadeghi et al. \citep{sadeghi2009hands} introduced a method that uses eye gaze information for interactive user-guided segmentation. Their approach involves three main steps: tracking eye gaze for seed pixel selection, using an optimization technique for image labeling based on user input, and integrating these elements to provide real-time visual feedback. This method has demonstrated significant speed-up in the process compared to traditional mouse controls. Besides, Gaze2Segment \citep{budd2021survey} combined biological and computer vision techniques to improve radiologists' diagnostic workflows, highlighting the potential of eye tracking in medical image analysis.

\textbf{Medical segmentation.} Medical segmentation is an essential and critical process for the identification, delineation, and visualization of regions of interest in medical images \citep{guo2018review}. This procedure is vital for accurate diagnosis, treatment planning, and monitoring of diseases. Recent advancements in deep learning methods, along with user-guided approaches such as eye tracking, have significantly improved the interactivity and responsiveness of segmentation tools in the medical field \citep{khosravan2017gaze2segment, sadeghi2009hands, budd2021survey}.

To the best of our knowledge, there is currently only one gaze-assisted interactive medical segmentation model, which was presented recently \cite{wang2023gazesam}. In this study, the authors used the original SAM to predict a segmentation mask based on a single gaze point. The authors compared gaze-based and mouse-click segmentation approaches and demonstrated that gaze-based segmentation was approximately two times faster and had slightly worse quality than click-based segmentation. However, they did not explore the implications of using multiple gaze points. So, it is quite unclear what happens if the gaze leaves the region of interest. This phenomenon may occur due to natural gaze variance. Furthermore, the study did not address how users should proceed if they need to remove a region from a segmentation mask rather than add one. By highlighting these gaps, our study aims to further investigate the potential of gaze data in the interactive segmentation of medical images.
\section{Methodology}

In this section, we introduce a novel method for gaze-assisted interactive segmentation of medical images. We adapt MedSAM to 1) a gaze point sequence of arbitrary length, 2) the fixed label (foreground) for prompt points, and 3) the natural variance of the gaze data. Thus, during the segmentation process, gaze points are automatically transmitted to the selected model, allowing real-time corrections to the segmentation mask generated by the base model.

\subsection{Problem definition}

A 3D medical image refers to volumetric data that represents anatomical structures in three dimensions, typically acquired through imaging modalities such as CT or MRI. We validated the segmentation adjustment with gaze using 3D CT images, where each image $\boldsymbol{X} \in \mathbb{R}^{H \times W \times D}$ consists of $D$ slices of shape $H \times W$. Thus, we transform CT scans to 2D slices for training pipeline and detailed analysis.

Medical segmentation aims to partition a medical image into meaningful regions, enabling the identification of anatomical structures and abnormalities. In the case of 3D CT, each image $\boldsymbol{X}$ is associated with mask $\boldsymbol{Y} \in \{0, 1\} ^ {H \times W \times D}$, where 1 corresponds to pixels containing the substance of interest, e.g. pancreas in abdominal CT, and 0 -- to background points. 

In this paper, we focus on interactive segmentation that leverages user interaction to refine initial segmentation masks. Specifically, we aim to utilize users' gaze to subsequently adjust the masks. Eye gaze data consists of gaze coordinates $(x_t, y_t)$ for the currently viewed slice and are available for each timestamp $t$. Our objective is to improve the segmentation mask in real-time using a sequence of gaze points $G_T=\{(x_t, y_t)\}_{t=0}^T$,
\begin{equation*}
    Q(M(X, G_T)) > Q(M(X, G_{T'})),
\end{equation*}
where $Q$ is a quality metric (higher is better, to be specific), $M$ is a gaze-assisted segmentation model generating mask $\boldsymbol{Y}$, and $G_{T'}$, where $T' < T$, is a previous sequence of gaze points. This formulation emphasizes the aim to improve segmentation quality by incorporating real-time gaze data.

\subsection{Model training with eye gaze coordinates}

\begin{figure*}[ht]
  \centering
   \includegraphics[width=0.9\linewidth]{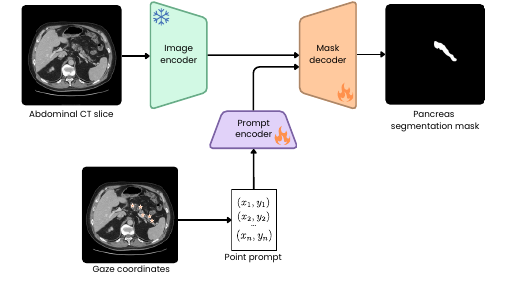}
   \caption{The proposed framework for gaze-assisted interactive segmentation of medical images. An illustrative example demonstrates the segmentation mask for the pancreas organ, which is predicted based on input gaze coordinates serving as a point prompt for the MedSAM model.}
   \label{fig:framework}
\end{figure*}

We introduce an approach for gaze-assisted interactive segmentation of medical images, which is depicted in Fig.~\ref{fig:framework}. The proposed framework is designed to integrate eye gaze coordinates into the segmentation process. To achieve this, we trained the MedSAM model, specifically the mask decoder and the prompt encoder components. We generate synthetic gaze points, simulating eye gaze coordinates input prompt of arbitrary length. In this setup, the input gaze data acts as a prompt sent to the MedSAM model's prompt encoder, guiding the segmentation process.

\textbf{Prompt generation from gaze data.} Initially, we attempted to fine-tune the MedSAM model using point prompts generated only from the reference segmentation mask. Nevertheless, this method led to poor accuracy during inference. This can be explained by the fact that training with points from the ground truth mask made the model overly sensitive to areas outside the region of interest, which is not plausible for eye gaze streams. To address this issue, we utilize synthetic eye gaze coordinates to create a more realistic representation of gaze behaviour. In such a method, 80\% of the generated points are chosen within the selected organ structure, specifically within the reference segmentation mask, while the remaining 20\% are positioned outside of this mask. We find it quite reasonable to assume that, during the segmentation process, the majority of gaze points will accurately fall within the region of interest (the anatomical structure). However, considering that some gaze points may accidentally fall outside the target area, we allocate the remaining 20\% of points outside the ground truth mask.

Secondly, we examine the dynamics of eye gaze movements, which are characterized by a series of rapid jumps or high-velocity movements, known as saccades, followed by fixations, which are located at periods in which the eye is stationary and remains relatively still \citep{blignaut2008effect}. By simulating these movements, we can generate random points that accurately reflect realistic gaze patterns. Alternatively, we can select random gaze coordinates, introducing some variability in the generated points. The latter approach is selected for generating point coordinates in prompts, as we believe it better reflects the natural fluctuations in gaze behaviour during real-world interactions. 

\begin{figure*}[ht]
  \centering
   \includegraphics[width=0.9\linewidth]{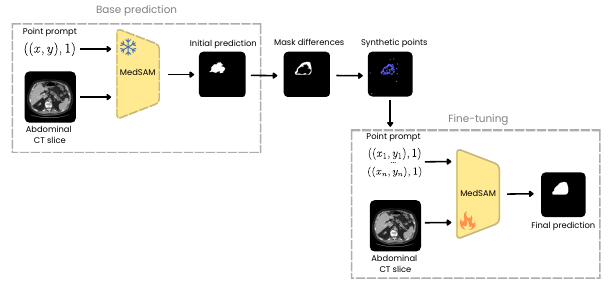}
   \caption{The provided pipeline outlines the steps for training a gaze-assisted segmentation model using synthetic gaze data points generated based on the mask difference approach. The process begins with the initial prediction from the frozen MedSAM. Using this initial prediction, we generate points that indicate the differences between the predicted mask and the ground truth, referred to as mask correction points. Next, the points are input into the same model to produce the final prediction. }
   \label{fig:mask_diff_framework}
\end{figure*}

Additionally, we propose another approach to generating point prompts based on mask differences. By analyzing the initial or base prediction, we create synthetic points that reflect the difference between the base prediction and the ground truth mask. In this approach, we allocate 70\% of the points within the mask difference, 20\% within the ground truth mask, and 10\% outside the ground truth. However, these proportions warrant a more detailed exploration in future work to optimize the strategy for point allocation. We subsequently train the model using these generated points (refer to Fig.~\ref{fig:mask_diff_framework}). These varied strategies enable the model to adapt to diverse gaze behaviours, thereby enhancing its segmentation capabilities through the incorporation of different input scenarios.

\textbf{Fine-tuning MedSAM.} As shown in Fig.~\ref{fig:framework}, we use a typical encoder-decoder architecture of MedSAM for interactive segmentation with a prompt encoder in the bottleneck. The difference between gaze and other prompt sources (for example, mouse clicks or bounding boxes) is 1) the inability to manually indicate which gaze points correspond to the user's desire to add/remove/keep unchanged the segmentation masks, 2) the inability to decline segmentation changes, 3) imprecision of the gaze data.  

While eye gaze data is generated as a sequence of coordinates for the fine-tuning pipeline, the SAM model requires the assignment of labels to input point coordinates, categorizing them as foreground (label 1), background (label 0), or not a point (label -1). However, justifying the inclusion or exclusion of specific points during experiments with an eye tracker poses challenges. To address this, we explored several labeling strategies for the point prompts.

Our strategies included: 1) randomly assigning points as either foreground or background, 2) removing point embedding and passing gaze coordinates to the prompt encoder without any associated label, and 3) assigning fixed labels (ones) to points and treating all gaze coordinates as foreground points. Ultimately, we found that all considered labeling strategies resulted in comparable performance. Following this, we choose to implement the latter, as it provides consistency for our experiments. We fine-tune the MedSAM model partially, specifically the prompt encoder and mask decoder, while keeping the other model's components frozen. 

\textbf{Interactive fine-tuning.} The inspiration for the suggested approach is based on two facts: 1) network probability map can be used as a confidence of prediction; 2) gaze data provides the viewer's region of interest. Combining these facts allows us to create an approach to interactive gaze-assisted segmentation. As illustrated in Fig.~\ref{fig:mask_diff_framework}, this methodology involves training the gaze points prompt encoder through simulations of human gaze behaviour.

Initially, a base prediction is generated using the point prompt within the ground truth mask. Subsequently, coordinates are derived based on the differences between the base predicted mask and the ground truth. These synthetic points are then utilized as prompts for the MedSAM prompt encoder, with gaze coordinates assigned a label "foreground" passed into the encoder. This approach is predicated on the assumption that it effectively mimics authentic gaze behaviour. Additionally, we introduce a similar strategy that extends beyond the utilization of point prompts generated from mask differences. This involves incorporating the model's base prediction mask as an additional input to the prompt encoder, along with the generated points. During inference, the previous mask can be employed with the eye gaze points, thereby enhancing the robustness of the segmentation process. 

\subsection{Interactive segmentation} 

Our interactive segmentation method provides users with visualization of predicted segmentation masks that they can adjust base predictions using gaze. This iterative approach allows for precise corrections in areas where automated segmentation may have fallen short. In our experiments, we will evaluate the effectiveness of the proposed approach by asking an expert to correct segmentation masks for various anatomical structures, stopping the correction when the mask is close to ideal.

\textbf{Radiology workstation.} We developed an emulation of a radiologist workstation with eye-tracking capabilities. The workstation was equipped with a lightweight and easy-to-use eye-tracking device. The developed software sequentially showed medical images, recorded gaze points with timestamps, and communicated with an interactive gaze-assisted segmentation model to update the displayed segmentation mask.

\section{Implementation details}

\subsection{Data} 

In this paper, we used the WORD dataset \citep{luo2021word}, which consists of 150 CT scans covering 16 abdominal organs. These CT scans were converted into 2D slices. Each slice was preprocessed to ensure that the corresponding ground truth mask contained a single distinct anatomical structure. Therefore, each organ can be represented as multiple structures when applicable. Similarly, to evaluate our models, we exploited the official test set of WORD.

\subsection{Eye tracking hardware} 

We developed a framework that simulates radiological workstations for CT slice analysis, enhanced with eye-tracking capabilities. This workstation was set up in an isolated room at our institution and features an LG diagnostic 10-bit monitor with a resolution of 3840×2160 pixels and a pixel density of 7.21 px/mm, along with a Tobii Eye Tracker 4C that operates at a frequency of 90 Hz.

Calibration of the eye tracker is essential prior to experimentation to ensure accurate tracking of eye movements and alignment of gaze coordinates with the user's point of focus. Once calibrated, users can perform segmentation tasks using the software, with the entire process controlled via a single Enter button. The framework effectively captures and synchronizes eye movement data, predicted masks, and controller commands, enabling a thorough examination of gaze behaviour within a radiological context.

\subsection{Training details} 

All our models were trained and tested using NVIDIA GTX 3090 Ti GPU. We employed the AdamW optimizer. We decreased the learning rate by 2 every 5 epochs with no improvement in validation loss. The initial learning rate was set at $5e^{-5}$, and the models were trained for 200 epochs. Additionally, we added early stopping with a patience of 10 epochs based on validation loss. 

For the MedSAM model, we focused on training only the prompt encoder and mask decoder, while other components of the model architecture were kept frozen. Due to resource constraints, we fine-tuned our models on a random sample from the WORD dataset containing around 5\% of all slices. Besides, the number of points in the point prompt was a critical hyperparameter. In this work, we report evaluations with 20 points, 50 points, and a random number of points ranging from 1 to 20 in the prompt. To incorporate a random number of points in the prompt for each sample during both training and testing, we padded some point prompts with a non-point label (-1). 

\section{Experiments and results}

\subsection{Proof of concept: synthetic gaze coordinates}

We developed a system for conducting experiments on synthetic gaze data. Unlike model validation during training, this system allowed for iterative testing of the model, simulating human responses to the model's predictions. After each prediction of the segmentation mask, we randomly generated gaze coordinates within the reference segmentation mask, in the areas of discrepancy between the predicted and reference masks, and outside of these areas. The gaze points were generated from a uniform random sample within the targeted areas, ensuring that no point was selected more than once (without replacement).

\textbf{Strategies for selecting gaze data.} We examined three general strategies for selecting gaze data during model inference. The first strategy involves accumulating gaze data across all iterations. Since the model was trained with a fixed size of the gaze point prompt, at each iteration, we randomly selected $N$ gaze points from the entire accumulated history. The second strategy is based on a linear combination of the nearest points between neighboring iterations. At each iteration, we selected $N$ points. For each point in the current iteration, we identified the nearest point from the set of points used to predict the current segmentation mask. Subsequently, we computed a linear combination between them. The coefficients of the linear combination, $\alpha$ and $(\alpha-1)$ were used to amplify or attenuate the significance of information from previous iterations. The initial gaze points were utilized during the first iteration. The last strategy involves using a model trained with a random number of points in the prompt. At each iteration, we randomly selected a fixed number of gaze points and data obtained during that iteration, thereby increasing the number of gaze points in the prompt by a fixed amount.

To test the first strategy, we generated 20 gaze points with varying distributions within the ground truth (gt) segmentation mask, mask differences, and beyond these areas. Table \ref{table-synth}, provided in the Appendix, demonstrates the average Dice Similarity Coefficient (DSC) for strategies with different gaze distributions. The highest average DSC of 0.9 is achieved when all gaze points are located within the reference segmentation mask.

To test the second strategy with a linear combination of points, we fixed the distribution of points (20\% in the gt mask, 70\% in the region of differences between the predicted and gt masks, and 10\% outside both regions) and changed the coefficient $\alpha$ from 0.1 to 0.9 in the linear combination of points. The greatest DSC of 0.89 was obtained at $\alpha = 0.6$, while the lowest value of 0.86 was obtained at  $\alpha = 0.9$. 

\begin{table}[t]
  \caption{Comparison of segmentation performance (DSC (\%)) with state-of-the-art models.}
  \label{table-sota}
  \centering
\begin{tabular}{l|c|c|c|c}
\toprule
Method & nnUNetV2 \citep{isensee2021nnu} & ResUNet \citep{diakogiannis2020resunet} & MedSAM \citep{MedSAM} & Ours MedSAM \\ \hline
Liver & 96.19 ± 2.16 & \textbf{96.55 ± 0.89} & 92.15 ± 1.22 & 95.14 ± 0.65 \\
Spleen & 94.33 ± 7.72 & 95.26 ± 2.84 & 78.68 ± 0.00 & \textbf{95.58 ± 0.27} \\
Kidney (L) & 91.29 ± 18.15 & 95.63 ± 1.20 & 83.40 ± 0.00 & \textbf{95.73 ± 0.03} \\
Kidney (R) & 91.20 ± 17.22 & 95.84 ± 1.16 & 91.93 ± 0.00 & \textbf{95.89 ± 0.01} \\
Stomach & 91.12 ± 3.60 & 91.58 ± 2.86 & 91.61 ± 0.00 & \textbf{94.70 ± 0.49} \\
Gallbladder & 83.19 ± 12.22 & 82.83 ± 11.8 & 85.43 ± 0.00 & \textbf{90.63 ± 0.02} \\
Esophagus & 77.79 ± 13.51 & 77.17 ± 14.68 & \textbf{88.94 ± 5.61} & 87.11 ± 2.70 \\
Pancreas & 83.55 ± 5.87 & 83.56 ± 5.60 & 79.78 ± 0.00 & \textbf{89.23 ± 0.34} \\
Duodenum & 64.47 ± 15.87 & 66.67 ± 15.36 & 80.07 ± 0.00 & \textbf{86.53 ± 0.34} \\
Colon & 83.92 ± 8.45 & 83.57 ± 8.69 & 75.96 ± 0.00 & \textbf{89.62 ± 2.43} \\
Intestine & 86.83 ± 4.02 & 86.76 ± 3.56 & 84.00 ± 0.00 & \textbf{88.28 ± 3.28} \\
Adrenal & 70.0 ± 11.86 & 70.9 ± 10.12 & 62.74 ± 5.28 & \textbf{80.28 ± 1.41} \\
Rectum & 81.49 ± 7.37 & 82.16 ± 6.73 & \textbf{92.54 ± 2.33} & 90.34 ± 1.65 \\
Bladder & 90.15 ± 16.85 & 91.0 ± 13.5 & 86.74 ± 0.00 & \textbf{94.16 ± 0.83} \\
Head of Femur (L) & 93.28 ± 5.12 & \textbf{93.39 ± 5.11} & 70.17 ± 0.00 & 93.19 ± 1.49 \\
Head of Femur (R) & \textbf{93.93 ± 4.29} & 93.88 ± 4.30 & 63.28 ± 0.00 & 93.00 ± 1.60 \\ \hline
Mean & 85.80 ± 5.27 & 86.67 ± 4.81 & 81.71 ± 9.39 & \textbf{90.54 ± 5.20} \\ 
\bottomrule
\end{tabular}
\end{table}

Table \ref{table-synth-20}, provided in the Appendix, presents the results of synthetic experiments conducted over 5 iterations with a model trained with a random number of points in the prompt. Different distributions of gaze points were tested between the reference segmentation mask, mask differences, and the remaining part of the image. The highest average DSC was achieved with a distribution of 20\% of points in the area of the reference segmentation mask, 70\% in the area of mask differences, and 10\% outside these areas, with two points selected at each iteration.

\textbf{Benchmarking.} In this section, we validate the performance of our proposed approach by comparing our best model strategy with 2D contemporary models benchmarked on WORD. Our model was trained using 20 points in the prompt, 80\% of the point coordinates located within the reference mask and the remaining 20\% positioned outside of it. The model was subsequently tested using the same approach, with 20 generated points in the input prompt. Besides, we include the original MedSAM in our comparison, testing it with bounding boxes. The bounding boxes are generated using Otsu's method based on the gt masks. 

The results, as shown in Table \ref{table-sota}, indicate that our method consistently outperforms the original MedSAM with bboxes and other state-of-the-art 2D models across various abdominal organs. Notably, our approach demonstrates significant improvements in segmentation performance (DSC), particularly in challenging cases such as the duodenum and adrenal organs.

\subsection{Human evaluation}

For our experiments, we engaged a proxy radiologist. The experiment consisted of two parts. In the first part, the radiologist annotated a set of $X$ CT slices using the proposed gaze-based interactive segmentation method. In the second part, the same dataset was annotated using the bounding box approach introduced in the MedSAM study \cite{MedSAM}.

\begin{table}[ht]
  \caption{DSC in human experiments. The point prompt consists of 20 gaze points. The strategies of gaze data selection with random gaze points (p), fixation points with replacement (fix rep), fixation points with accumulation (fix accum), a linear combination of gaze points (linear), accumulation of 2 gaze points at each iteration (accum 2 p) are demonstrated. Additionally, the approach with segmentation using generated bounding boxes (MedSAM) is shown. }
  \label{table-20}
  \centering
\begin{tabular}{lllllll}
\toprule
organ            & points & fix rep & fix accum  & linear & accum 2 p & MedSAM \\
\midrule
Liver             & 0.942         & 0.626            & 0.655            & 0.888              & 0.958               & 0.967      \\
Spleen            & 0.931         & 0.685            & 0.871            & 0.642              & 0.915               & 0.972      \\
Kidney (L)        & 0.959         & 0.796            & 0.961            & 0.930              & 0.958               & 0.957      \\
Kidney (R)        & 0.848         & 0.545            & 0.884            & 0.895              & 0.758               & 0.878      \\
Stomach           & 0.870         & 0.828            & 0.951            & 0.944              & 0.906               & 0.941      \\
Gallbladder       & 0.909         & 0.896            & 0.906            & 0.903              & 0.882               & 0.908     \\
Esophagus         & 0.913         & 0.906            & 0.920            & 0.766              & 0.854               & 0.881     \\
Pancreas          & 0.877         & 0.528            & 0.901            & 0.882              & 0.915               & 0.916      \\
Duodenum          & 0.839         & 0.851            & 0.840            & 0.730              & 0.846               & 0.623      \\
Colon             & 0.885         & 0.869            & 0.873            & 0.773              & 0.851               & 0.916      \\
Intestine         & 0.786         & 0.729            & 0.778            & 0.833              & 0.797               & 0.801      \\
Adrenal           & 0.784         & 0.764            & 0.765            & 0.796              & 0.772               & 0.793      \\
Rectum            & 0.901         & 0.744            & 0.795            & 0.666              & 0.935               & 0.948      \\
Bladder           & 0.965         & 0.069            & 0.113            & 0.940              & 0.966               & 0.945      \\
Head of Femur (L) & 0.749         & 0.873            & 0.871            & 0.782              & 0.902               & 0.945      \\
Head of Femur (R) & 0.871         & 0.861            & 0.914            & 0.855              & 0.869               & 0.740      \\
\midrule
Mean              & 0.854         & 0.764            & 0.831            & 0.820              & 0.861               & 0.884     \\
\bottomrule
\end{tabular}
\end{table}

Furthermore, we conducted a series of experiments involving the proxy radiologist, employing various gaze data selection strategies and different fine-tuned models. For the model trained with a prompt consisting of 20 points, the results are presented in Table \ref{table-20}. The approach of selecting gaze data based on 20 randomly accumulated points achieved an average Dice coefficient of 0.854. In contrast, the method based on selecting 20 random fixation points with replacement demonstrated an average Dice coefficient of 0.764, whereas the method based on selecting random fixation points with accumulation achieved an average Dice coefficient of 0.831. The approach with a linear combination of random points resulted in an average Dice coefficient of 0.82. Meanwhile, the method employing a model trained with a random number of points in the prompt, accumulating 2 random gaze points at each iteration, achieved an average Dice coefficient of 0.861. The original MedSAM with a bounding box prompt obtained an average Dice coefficient of 0.884. However, for certain organs, such as the right and left kidneys and the duodenum, the performance of MedSAM with bboxes was comparatively lower. The average time for the gaze-assisted segmentation approach with various gaze point selection strategies was 9.7 ± 4.9 seconds, compared to 5.7 ± 3.1 seconds required for MedSAM with bboxes. We provide visualization results of interactive prompts along with predictions made based on the expert's eye gaze in the Appendix (see Figure~\ref{fig:vis-steps-9}).

\begin{table}[t]
  \caption{DSC in human experiments. The point prompt consists of 20 gaze points. The model with the segmentation mask from the previous iteration in the prompt was used. The strategies of gaze data selection with random gaze points (p), fixation points with replacement (fix rep), and fixation points with accumulation (fix accum) are shown.}
  \label{table-20-prev_mask}
  \centering
\begin{tabular}{llll}
\toprule
organ  & points & fix rep & fix accum \\
\midrule
Liver             & 0.661         & 0.253            & 0.640            \\
Spleen            & 0.863         & 0.644            & 0.919            \\
Kidney (L)        & 0.852         & 0.593            & 0.914            \\
Kidney (R)        & 0.756         & 0.391            & 0.765            \\
Stomach           & 0.715         & 0.729            & 0.751            \\
Gallbladder       & 0.895         & 0.713            & 0.886            \\
Esophagus         & 0.866         & 0.863            & 0.873            \\
Pancreas          & 0.873         & 0.245            & 0.862            \\
Duodenum          & 0.725         & 0.666            & 0.792            \\
Colon             & 0.773         & 0.577            & 0.740            \\
Intestine         & 0.657         & 0.638            & 0.671            \\
Adrenal           & 0.667         & 0.423            & 0.491            \\
Rectum            & 0.748         & 0.738            & 0.725            \\
Bladder           & 0.765         & 0.306            & 0.725            \\
Head of Femur (L) & 0.809         & 0.712            & 0.824            \\
Head of Femur (R) & 0.764         & 0.771            & 0.764            \\
\midrule
Mean              & 0.745         & 0.601            & 0.745   \\
\bottomrule
\end{tabular}
\end{table}

Additionally, we evaluated several gaze point selection strategies using a model trained on 20 points in the prompt along with the predicted segmentation mask from the previous iteration. Table \ref{table-20-prev_mask} presents the segmentation quality metrics for each organ using gaze-assisted segmentation. The approach employing the strategy of selecting gaze data based on 20 randomly accumulated points achieved an average Dice coefficient of 0.745, while the method based on selecting 20 random fixation points with replacement yielded an average Dice coefficient of 0.601. The method using random fixation points with accumulation also resulted in an average Dice coefficient of 0.745.

\subsection{Limitations}

Our work has several limitations. 1) Despite the large number of human experiments, only one participant was involved. Future work should include more participants. 2) The participant segmented organs on individual slices. Future work should focus on using 3-dimensional medical data with the ability to manipulate slices.
\section{Conclusion and future work}

Medical image analysis has always been challenging, leaving room for improvement. The proposed gaze-assisted approach for organ segmentation in abdominal CT scans achieves state-of-the-art performance. Moreover, we conducted a series of synthetic experiments to identify the best approach, which was used in the eye-tracking experiments with a proxy radiologist. We observed that the performance of gaze-assisted MedSAM was superior to alternative state-of-the-art solutions. 

Future work should expand the application of our gaze-assisted segmentation approach to a wider variety of medical datasets beyond abdominal imaging, including different anatomical regions and medical conditions. This will enhance the robustness and applicability of our method. We also intend to explore strategies for adapting our solution to 3D data, focusing on effective segmentation techniques for volumetric images. 
\section{Acknowledgement}

This research has been financially supported by The Analytical Center for the Government of the Russian Federation (Agreement No. 70-2021-00143 01.11.2021, IGK 000000D730324P540002).

\newpage


{
\small
\bibliography{main}

\begin{thebibliography}{10}

\bibitem{Alexander2022}
Robert Alexander, Stephen Waite, Michael~A. Bruno, Elizabeth~A. Krupinski, Leonard Berlin, Stephen Macknik, and Susana Martinez-Conde.
\newblock Mandating limits on workload, duty, and speed in radiology.
\newblock {\em Radiology}, 304(2):274--282, August 2022.

\bibitem{blignaut2008effect}
Pieter Blignaut and Tanya Beelders.
\newblock The effect of fixational eye movements on fixation identification with a dispersion-based fixation detection algorithm.
\newblock {\em Journal of eye movement research}, 2(5), 2008.

\bibitem{Borys2023}
Katarzyna Borys, Yasmin~Alyssa Schmitt, Meike Nauta, Christin Seifert, Nicole Kr\"{a}mer, Christoph~M. Friedrich, and Felix Nensa.
\newblock Explainable {AI} in medical imaging: An overview for clinical practitioners {\textendash} beyond saliency-based {XAI} approaches.
\newblock {\em European Journal of Radiology}, 162:110786, May 2023.

\bibitem{Boykov}
Y.Y. Boykov and M.-P. Jolly.
\newblock Interactive graph cuts for optimal boundary \& region segmentation of objects in n-d images.
\newblock In {\em Proceedings Eighth IEEE International Conference on Computer Vision. ICCV 2001}, volume~1, pages 105--112 vol.1, 2001.

\bibitem{Bruls2020}
R.~J.~M. Bruls and R.~M. Kwee.
\newblock Workload for radiologists during on-call hours: dramatic increase in the past 15~years.
\newblock {\em Insights into Imaging}, 11(1), November 2020.

\bibitem{budd2021survey}
Samuel Budd, Emma~C Robinson, and Bernhard Kainz.
\newblock A survey on active learning and human-in-the-loop deep learning for medical image analysis.
\newblock {\em Medical image analysis}, 71:102062, 2021.

\bibitem{Chen2022}
Haomin Chen, Catalina Gomez, Chien-Ming Huang, and Mathias Unberath.
\newblock Explainable medical imaging {AI} needs human-centered design: guidelines and evidence from a systematic review.
\newblock {\em npj Digital Medicine}, 5(1), October 2022.

\bibitem{chen2021deep}
Xuming Chen, Shanlin Sun, Narisu Bai, Kun Han, Qianqian Liu, Shengyu Yao, Hao Tang, Chupeng Zhang, Zhipeng Lu, Qian Huang, et~al.
\newblock A deep learning-based auto-segmentation system for organs-at-risk on whole-body computed tomography images for radiation therapy.
\newblock {\em Radiotherapy and Oncology}, 160:175--184, 2021.

\bibitem{diakogiannis2020resunet}
Foivos~I Diakogiannis, Fran{\c{c}}ois Waldner, Peter Caccetta, and Chen Wu.
\newblock Resunet-a: A deep learning framework for semantic segmentation of remotely sensed data.
\newblock {\em ISPRS Journal of Photogrammetry and Remote Sensing}, 162:94--114, 2020.

\bibitem{Gaube2021}
Susanne Gaube, Harini Suresh, Martina Raue, Alexander Merritt, Seth~J. Berkowitz, Eva Lermer, Joseph~F. Coughlin, John~V. Guttag, Errol Colak, and Marzyeh Ghassemi.
\newblock Do as {AI} say: susceptibility in deployment of clinical decision-aids.
\newblock {\em npj Digital Medicine}, 4(1), February 2021.

\bibitem{Grady2006}
L.~Grady.
\newblock Random walks for image segmentation.
\newblock {\em IEEE Transactions on Pattern Analysis and Machine Intelligence}, 28(11):1768--1783, 2006.

\bibitem{guo2018review}
Yanming Guo, Yu~Liu, Theodoros Georgiou, and Michael~S Lew.
\newblock A review of semantic segmentation using deep neural networks.
\newblock {\em International journal of multimedia information retrieval}, 7:87--93, 2018.

\bibitem{isensee2021nnu}
Fabian Isensee, Paul~F Jaeger, Simon~AA Kohl, Jens Petersen, and Klaus~H Maier-Hein.
\newblock nnu-net: a self-configuring method for deep learning-based biomedical image segmentation.
\newblock {\em Nature methods}, 18(2):203--211, 2021.

\bibitem{Karargyris2021}
Alexandros Karargyris, Satyananda Kashyap, Ismini Lourentzou, Joy~T. Wu, Arjun Sharma, Matthew Tong, Shafiq Abedin, David Beymer, Vandana Mukherjee, Elizabeth~A. Krupinski, and Mehdi Moradi.
\newblock Creation and validation of a chest x-ray dataset with eye-tracking and report dictation for {AI} development.
\newblock {\em Scientific Data}, 8(1), March 2021.

\bibitem{khosravan2017gaze2segment}
Naji Khosravan, Haydar Celik, Baris Turkbey, Ruida Cheng, Evan McCreedy, Matthew McAuliffe, Sandra Bednarova, Elizabeth Jones, Xinjian Chen, Peter Choyke, et~al.
\newblock Gaze2segment: a pilot study for integrating eye-tracking technology into medical image segmentation.
\newblock In {\em Medical Computer Vision and Bayesian and Graphical Models for Biomedical Imaging: MICCAI 2016 International Workshops, MCV and BAMBI, Athens, Greece, October 21, 2016, Revised Selected Papers 8}, pages 94--104. Springer, 2017.

\bibitem{sam2023}
Alexander Kirillov, Eric Mintun, Nikhila Ravi, Hanzi Mao, Chloe Rolland, Laura Gustafson, Tete Xiao, Spencer Whitehead, Alexander~C Berg, Wan-Yen Lo, et~al.
\newblock Segment anything.
\newblock In {\em Proceedings of the IEEE/CVF International Conference on Computer Vision}, pages 4015--4026, 2023.

\bibitem{DanLantsman2022}
Christine~Dan Lantsman, Yiftach Barash, Eyal Klang, Larisa Guranda, Eli Konen, and Noam Tau.
\newblock Trend in radiologist workload compared to number of admissions in the emergency department.
\newblock {\em European Journal of Radiology}, 149:110195, April 2022.

\bibitem{Li2022}
Guangqi Li, Xin Wu, and Xuelei Ma.
\newblock Artificial intelligence in radiotherapy.
\newblock {\em Seminars in Cancer Biology}, 86:160--171, November 2022.

\bibitem{Li2023}
Matthew~D. Li and Brent~P. Little.
\newblock Appropriate reliance on artificial intelligence in radiology education.
\newblock {\em Journal of the American College of Radiology}, June 2023.

\bibitem{luo2021word}
Xiangde Luo, Wenjun Liao, Jianghong Xiao, Jieneng Chen, Tao Song, Xiaofan Zhang, Kang Li, Dimitris~N Metaxas, Guotai Wang, and Shaoting Zhang.
\newblock Word: A large scale dataset, benchmark and clinical applicable study for abdominal organ segmentation from ct image.
\newblock {\em arXiv preprint arXiv:2111.02403}, 2021.

\bibitem{MedSAM}
Jun Ma, Yuting He, Feifei Li, Lin Han, Chenyu You, and Bo~Wang.
\newblock Segment anything in medical images.
\newblock {\em Nature Communications}, 15(1):654, 2024.

\bibitem{orig_sam_for_medical}
Maciej~A. Mazurowski, Haoyu Dong, Hanxue Gu, Jichen Yang, Nicholas Konz, and Yixin Zhang.
\newblock Segment anything model for medical image analysis: an experimental study.
\newblock {\em Medical Image Analysis}, 2023.

\bibitem{McGrath2022}
Anika~L. McGrath, Katerina Dodelzon, Omer~A. Awan, Nicholas Said, and Puneet Bhargava.
\newblock Optimizing radiologist productivity and efficiency: Work smarter, not harder.
\newblock {\em European Journal of Radiology}, 155:110131, October 2022.

\bibitem{Pershin2022}
Ilya Pershin, Maksim Kholiavchenko, Bulat Maksudov, Tamerlan Mustafaev, Dilyara Ibragimova, and Bulat Ibragimov.
\newblock Artificial intelligence for the analysis of workload-related changes in radiologists' gaze patterns.
\newblock {\em {IEEE} Journal of Biomedical and Health Informatics}, 26(9):4541--4550, September 2022.

\bibitem{Pershin2023}
Ilya Pershin, Tamerlan Mustafaev, Dilyara Ibragimova, and Bulat Ibragimov.
\newblock Changes in radiologists' gaze patterns against lung x-rays with different abnormalities: a randomized experiment.
\newblock {\em Journal of Digital Imaging}, 36(3):767--775, January 2023.

\bibitem{Reyes2020}
Mauricio Reyes, Raphael Meier, S{\'{e}}rgio Pereira, Carlos~A. Silva, Fried-Michael Dahlweid, Hendrik von Tengg-Kobligk, Ronald~M. Summers, and Roland Wiest.
\newblock On the interpretability of artificial intelligence in radiology: Challenges and opportunities.
\newblock {\em Radiology: Artificial Intelligence}, 2(3):e190043, May 2020.

\bibitem{sadeghi2009hands}
Maryam Sadeghi, G~Tien, Ghassan Hamarneh, and Margaret Atkins.
\newblock Hands-free interactive image segmentation using eyegaze.
\newblock {\em Progress in Biomedical Optics and Imaging - Proceedings of SPIE}, 7260, 02 2009.

\bibitem{sharma2010automated}
Neeraj Sharma and Lalit~M Aggarwal.
\newblock Automated medical image segmentation techniques.
\newblock {\em Journal of medical physics}, 35(1):3--14, 2010.

\bibitem{tahri2023eye}
Mohammed Tahri~Sqalli, Begali Aslonov, Mukhammadjon Gafurov, Nurmukhammad Mukhammadiev, and Yahya Sqalli~Houssaini.
\newblock Eye tracking technology in medical practice: a perspective on its diverse applications.
\newblock {\em Frontiers in Medical Technology}, 5:1253001, 2023.

\bibitem{TaylorPhillips2019}
Sian Taylor-Phillips and Chris Stinton.
\newblock Fatigue in radiology: a fertile area for future research.
\newblock {\em The British Journal of Radiology}, 92(1099):20190043, July 2019.

\bibitem{wang2023gazesam}
Bin Wang, Armstrong Aboah, Zheyuan Zhang, Hongyi Pan, and Ulas Bagci.
\newblock Gazesam: Interactive image segmentation with eye gaze and segment anything model.
\newblock In {\em NeuRIPS 2023 Workshop on Gaze Meets ML}, 2023.

\bibitem{Wang2022}
Sheng Wang, Xi~Ouyang, Tianming Liu, Qian Wang, and Dinggang Shen.
\newblock Follow my eye: Using gaze to supervise computer-aided diagnosis.
\newblock {\em {IEEE} Transactions on Medical Imaging}, 41(7):1688--1698, July 2022.

\bibitem{medsam_review}
Chunhui Zhang, Li~Liu, Yawen Cui, Guanjie Huang, Weilin Lin, Yiqian Yang, and Yuehong Hu.
\newblock A comprehensive survey on segment anything model for vision and beyond, 2023.

\bibitem{zhang2023segment}
Lian Zhang, Zhengliang Liu, Lu~Zhang, Zihao Wu, Xiaowei Yu, Jason Holmes, Hongying Feng, Haixing Dai, Xiang Li, Quanzheng Li, et~al.
\newblock Segment anything model (sam) for radiation oncology.
\newblock {\em arXiv preprint arXiv:2306.11730}, 2023.

\bibitem{seem}
Xueyan Zou, Jianwei Yang, Hao Zhang, Feng Li, Linjie Li, Jianfeng Wang, Lijuan Wang, Jianfeng Gao, and Yong~Jae Lee.
\newblock Segment everything everywhere all at once.
\newblock {\em Advances in Neural Information Processing Systems}, 36, 2024.

\end{thebibliography}
}

\newpage
\appendix
\section{Appendix / supplemental material}

\begin{table}[ht]
  \caption{Comparison of segmentation performance (DSC) in synthetic experiments with accumulation-based gaze data selection strategy with the distribution of points within the ground truth mask (prop gt), in the region of differences between the predicted and ground truth masks (prop mask diff) and outside both regions (prop out).}
  \label{table-synth}
  \centering
\begin{tabular}{llll}
\toprule
prop gt& prop out    & prop mask diff      & DSC             \\
\midrule
0\%                     & 100\%        & \multirow{12}{*}{0\%} & 0.07115          \\
10\%                    & 90\%         &                       & 0.20640          \\
20\%                    & 80\%         &                       & 0.34013          \\
30\%                    & 70\%         &                       & 0.52645          \\
40\%                    & 60\%         &                       & 0.74141          \\
50\%                    & 50\%         &                       & 0.82396          \\
60\%                    & 40\%         &                       & 0.85157          \\
70\%                    & 30\%         &                       & 0.87200          \\
80\%                    & 20\%         &                       & 0.88210          \\
90\%                    & 10\%         &                       & 0.89290          \\
95\%                    & 5\%          &                       & 0.89940          \\
\textbf{100\%}          & \textbf{0\%} &                       & \textbf{0.90010} \\
\midrule
20\%                    & 10\%         & 70\%                  & 0.82960          \\
5\%                     & 5\%          & 90\%                  & 0.80380          \\
40\%                    & 10\%         & 50\%                  & 0.85080          \\
30\%                    & 10\%         & 60\%                  & 0.84320          \\
60\%                    & 10\%         & 30\%                  & 0.86662          \\
70\%                    & 10\%         & 20\%                  & 0.87379          \\
80\%                    & 5\%          & 15\%                  & 0.87848          \\
90\%                    & 5\%          & 5\%                   & 0.87930       \\  
\bottomrule
\end{tabular}
\end{table}

\begin{table}[ht]
  \caption{Comparison of segmentation performance (DSC) in synthetic experiments with gaze data selection strategy based on sequential accumulation of gaze data in the prompt at each iteration with the distribution of points in the gt mask (prop gt), in the region of differences between the predicted and reference masks (prop mask diff) and outside both regions (prop out). For each case, a fixed number of points (num points) was selected at each iteration.}
  \label{table-synth-20}
  \centering
\begin{tabular}{lllll}
\toprule
prop gt & prop out     &  prop mask diff & num points                 & dice             \\
\midrule
20\%                    & 10\%          & 70\%             & \multirow{8}{*}{1}          & 0.84142          \\
5\%                     & 5\%           & 90\%             &                             & 0.83411          \\
40\%                    & 10\%          & 50\%             &                             & 0.82446          \\
30\%                    & 10\%          & 60\%             &                             & 0.80465          \\
60\%                    & 10\%          & 30\%             &                             & 0.75430          \\
70\%                    & 10\%          & 20\%             &                             & 0.77780          \\
80\%                    & 5\%           & 15\%             &                             & 0.84267          \\
90\%                    & 5\%           & 5\%              &                             & 0.82022          \\
\midrule
\textbf{20\%}           & \textbf{10\%} & \textbf{70\%}    & \multirow{8}{*}{\textbf{2}} & \textbf{0.88769} \\
5\%                     & 5\%           & 90\%             &                             & 0.77403          \\
40\%                    & 10\%          & 50\%             &                             & 0.75456          \\
30\%                    & 10\%          & 60\%             &                             & 0.82651          \\
60\%                    & 10\%          & 30\%             &                             & 0.86370          \\
70\%                    & 10\%          & 20\%             &                             & 0.81802          \\
80\%                    & 5\%           & 15\%             &                             & 0.83697          \\
90\%                    & 5\%           & 5\%              &                             & 0.79542          \\
\midrule
20\%                    & 10\%          & 70\%             & 
\multirow{8}{*}{4}          & 0.67183          \\
5\%                     & 5\%           & 90\%             &                             & 0.70947          \\
40\%                    & 10\%          & 50\%             &                             & 0.80588          \\
30\%                    & 10\%          & 60\%             &                             & 0.80906          \\
60\%                    & 10\%          & 30\%             &                             & 0.88311          \\
70\%                    & 10\%          & 20\%             &                             & 0.86640          \\
80\%                    & 5\%           & 15\%             &                             & 0.85710          \\
90\%                    & 5\%           & 5\%              &                             & 0.87252          \\
\midrule
20\%                    & 10\%          & 70\%             & \multirow{8}{*}{5}          & 0.85304          \\

5\%                     & 5\%           & 90\%             &                             & 0.87967          \\
40\%                    & 10\%          & 50\%             &                             & 0.87501          \\
30\%                    & 10\%          & 60\%             &                             & 0.86181          \\
60\%                    & 10\%          & 30\%             &                             & 0.84695          \\
70\%                    & 10\%          & 20\%             &                             & 0.83883          \\
80\%                    & 5\%           & 15\%             &                             & 0.88035          \\
90\%                    & 5\%           & 5\%              &                             & 0.86934   \\  
\bottomrule
\end{tabular}
\end{table}

\begin{figure*}[ht]
  \centering
  
  \begin{subfigure}{0.3\linewidth}
    \includegraphics[width=\linewidth]{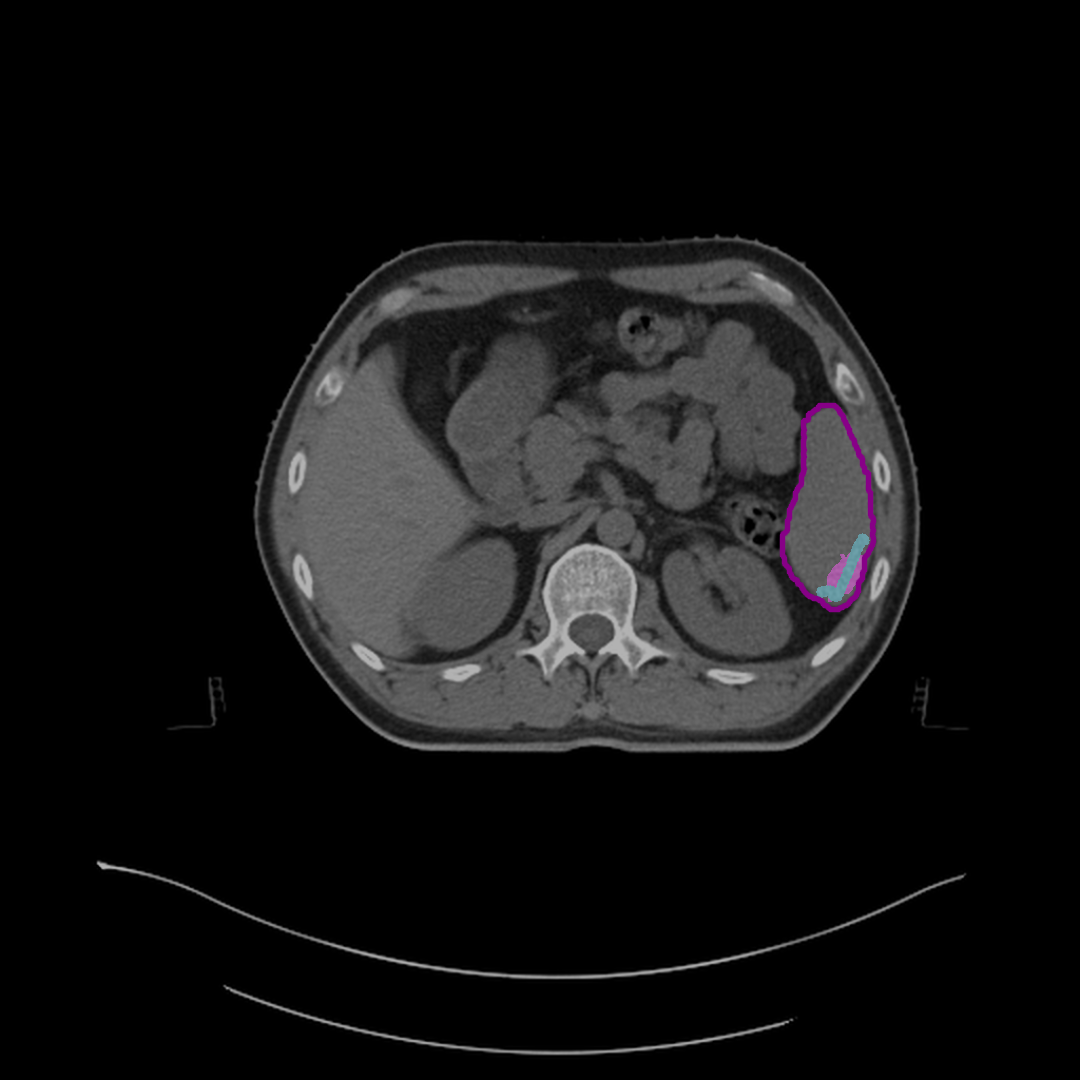}
    \caption{Initial predicted mask.}
    \label{fig:short-a-ap}
  \end{subfigure}
  \hfill
  \begin{subfigure}{0.3\linewidth}
    \includegraphics[width=\linewidth]{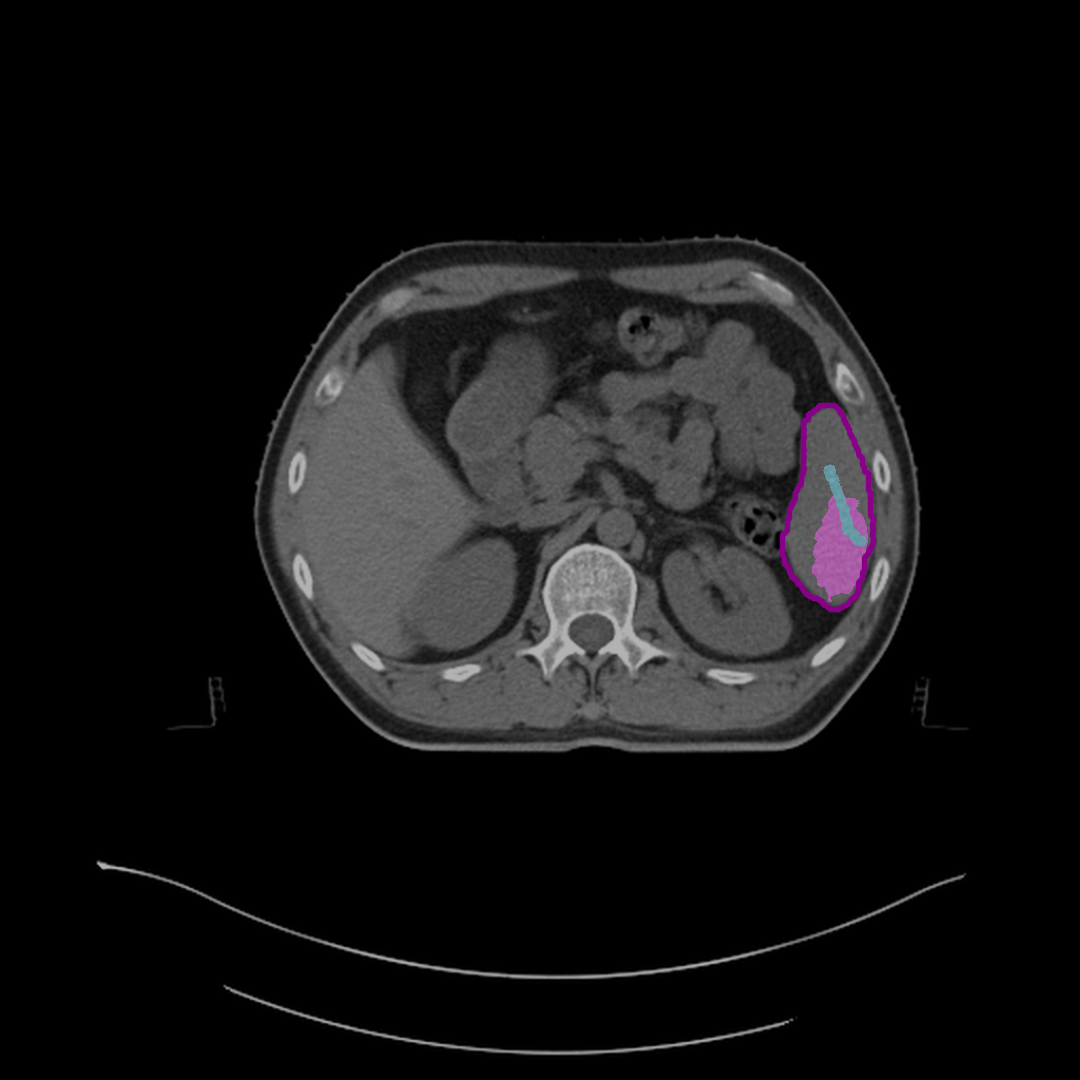}
    \caption{Mask after half correction.}
    \label{fig:short-b-ap}
  \end{subfigure}
  \hfill
  \begin{subfigure}{0.3\linewidth}
  \includegraphics[width=\linewidth]{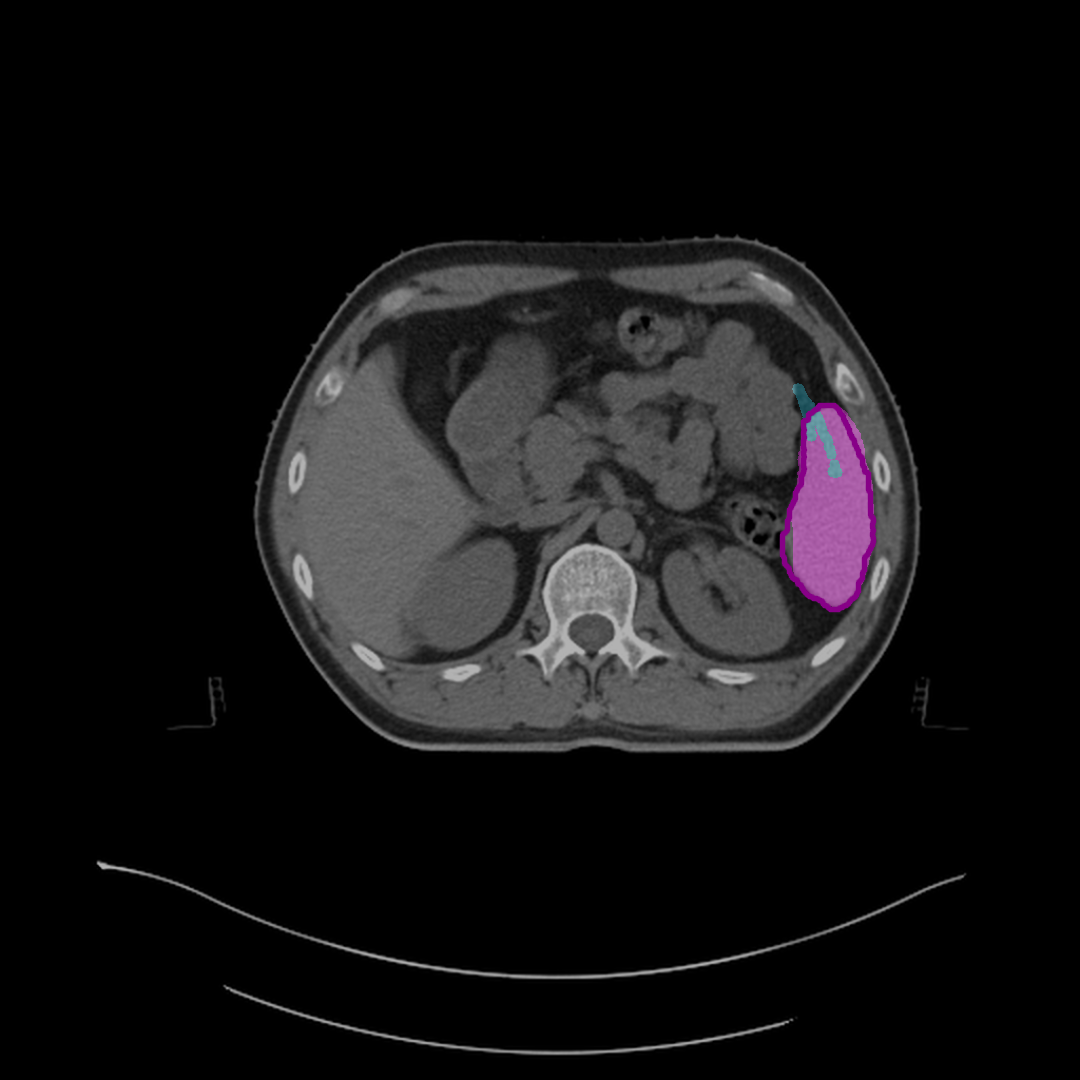}
  \caption{Mask after full correction.}
  \label{fig:short-c-ap}
  \end{subfigure}

  \begin{subfigure}{0.3\linewidth}
    \includegraphics[width=\linewidth]{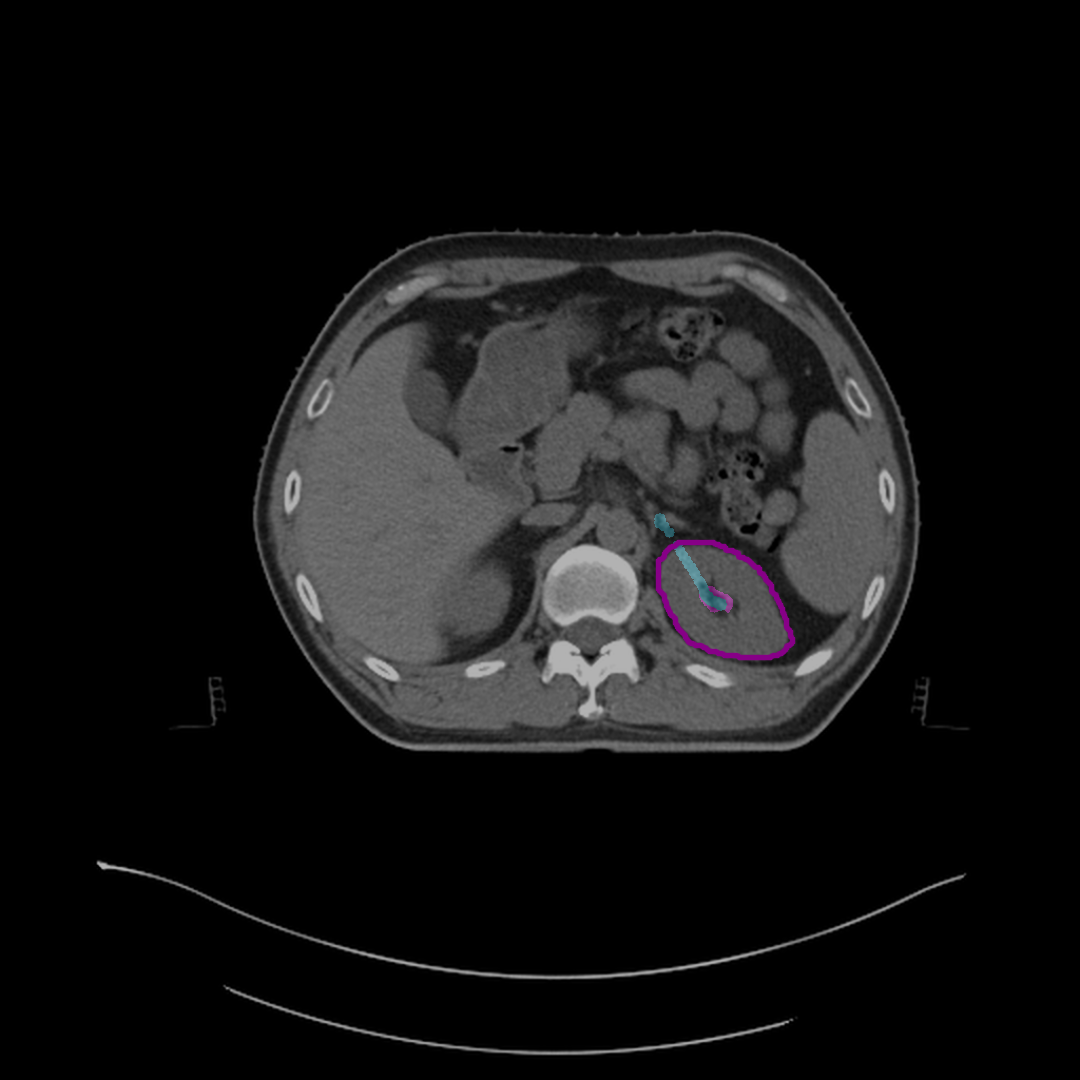}
    \caption{Initial predicted mask.}
    \label{fig:short-d-ap}
  \end{subfigure}
  \hfill
  \begin{subfigure}{0.3\linewidth}
    \includegraphics[width=\linewidth]{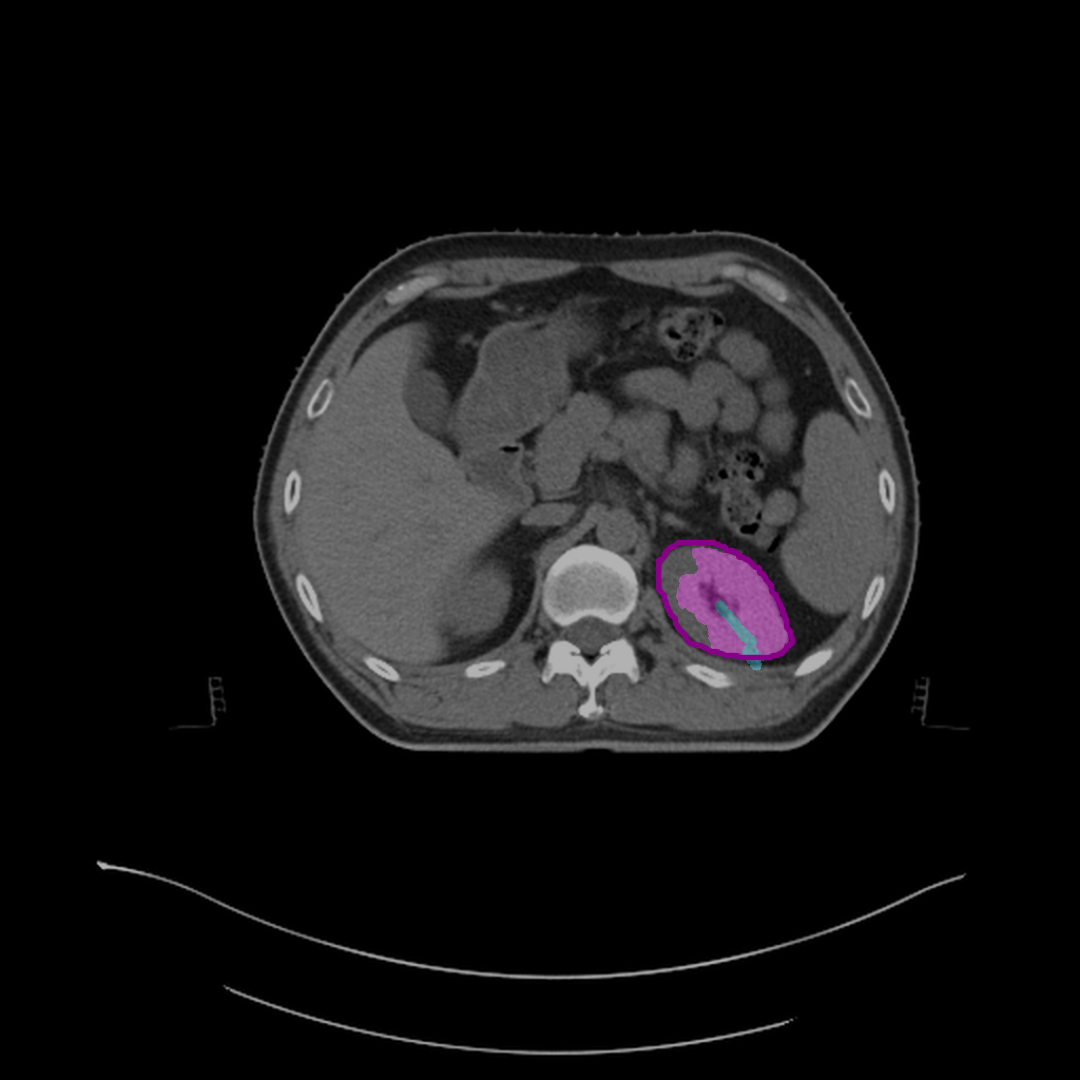}
    \caption{Mask after half correction.}
    \label{fig:short-e-ap}
  \end{subfigure}
  \hfill
  \begin{subfigure}{0.3\linewidth}
  \includegraphics[width=\linewidth]{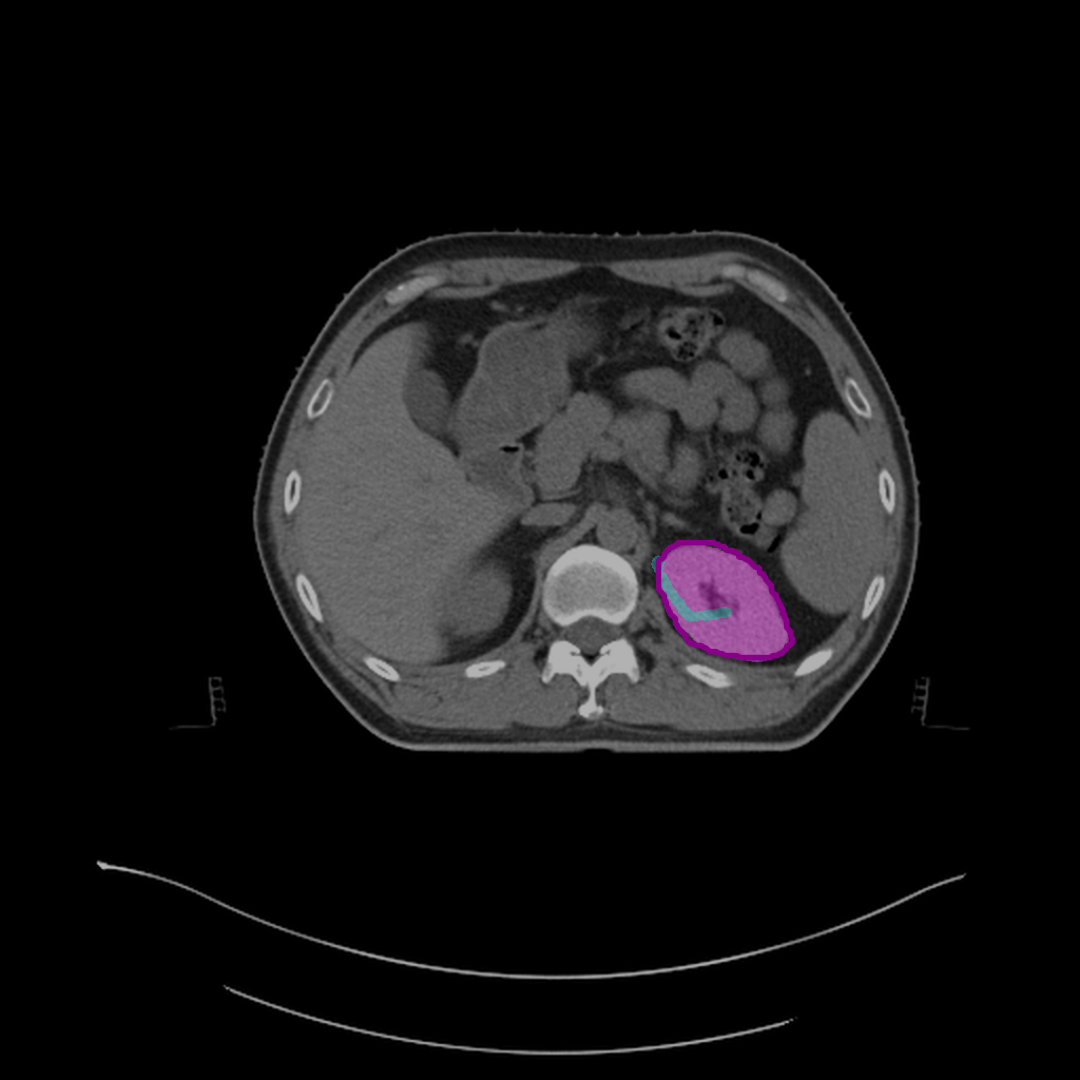}
  \caption{Mask after full correction.}
  \label{fig:short-f-ap}
  \end{subfigure}

  \begin{subfigure}{0.3\linewidth}
    \includegraphics[width=\linewidth]{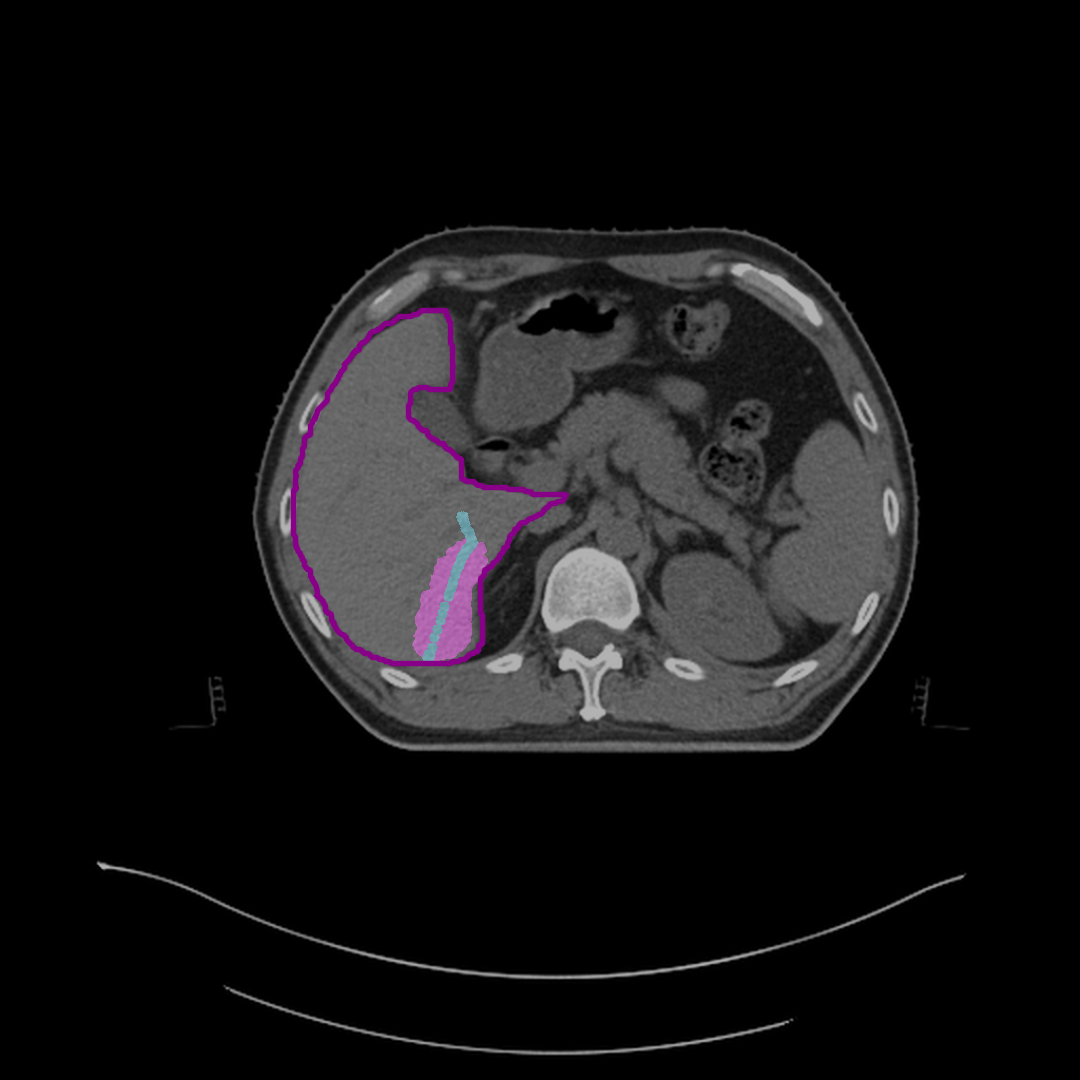}
    \caption{Initial predicted mask.}
    \label{fig:short-g-ap}
  \end{subfigure}
  \hfill  
  \begin{subfigure}{0.3\linewidth}
    \includegraphics[width=\linewidth]{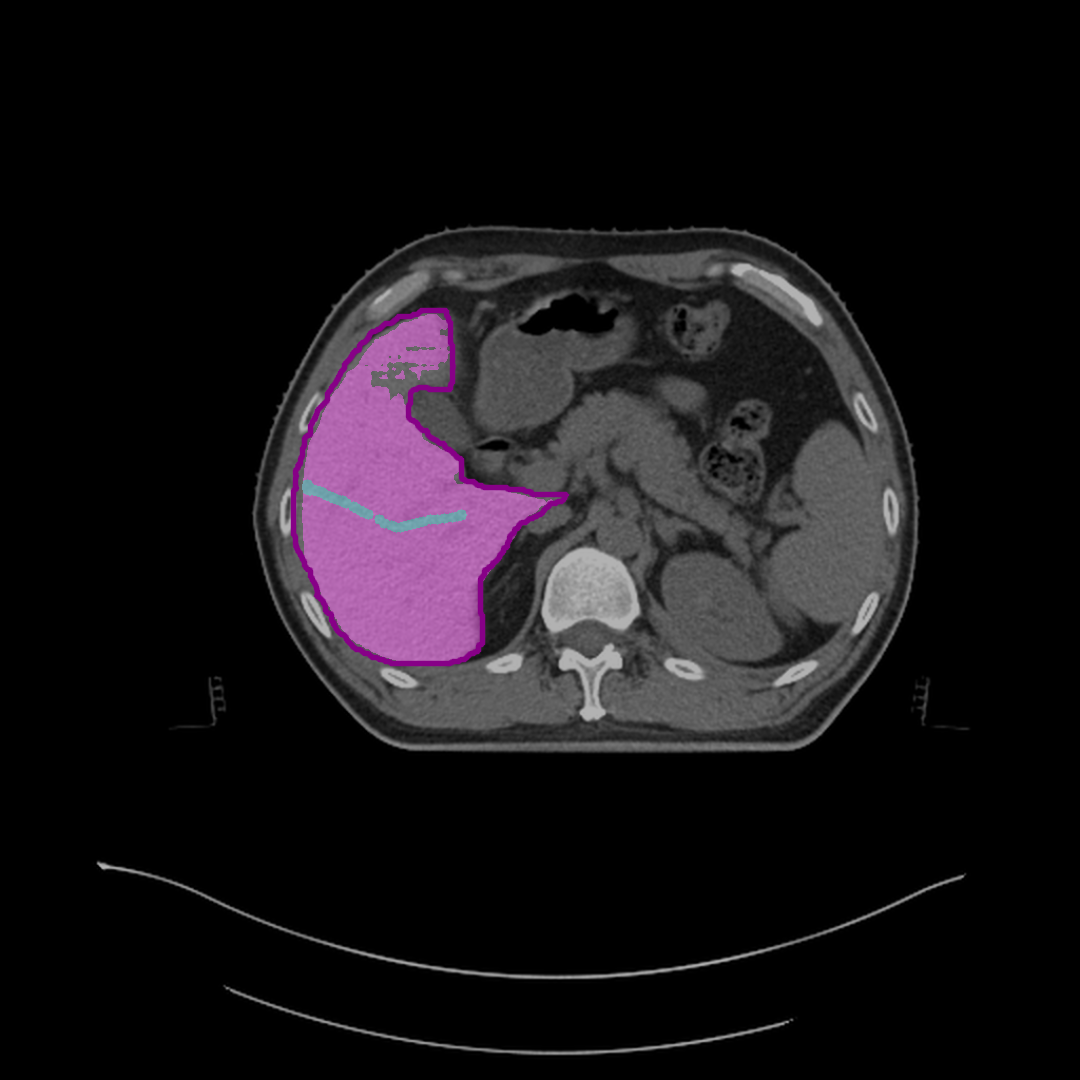}
    \caption{Mask after half correction.}
    \label{fig:short-h-ap}
  \end{subfigure}
  \hfill
  \begin{subfigure}{0.3\linewidth}
  \includegraphics[width=\linewidth]{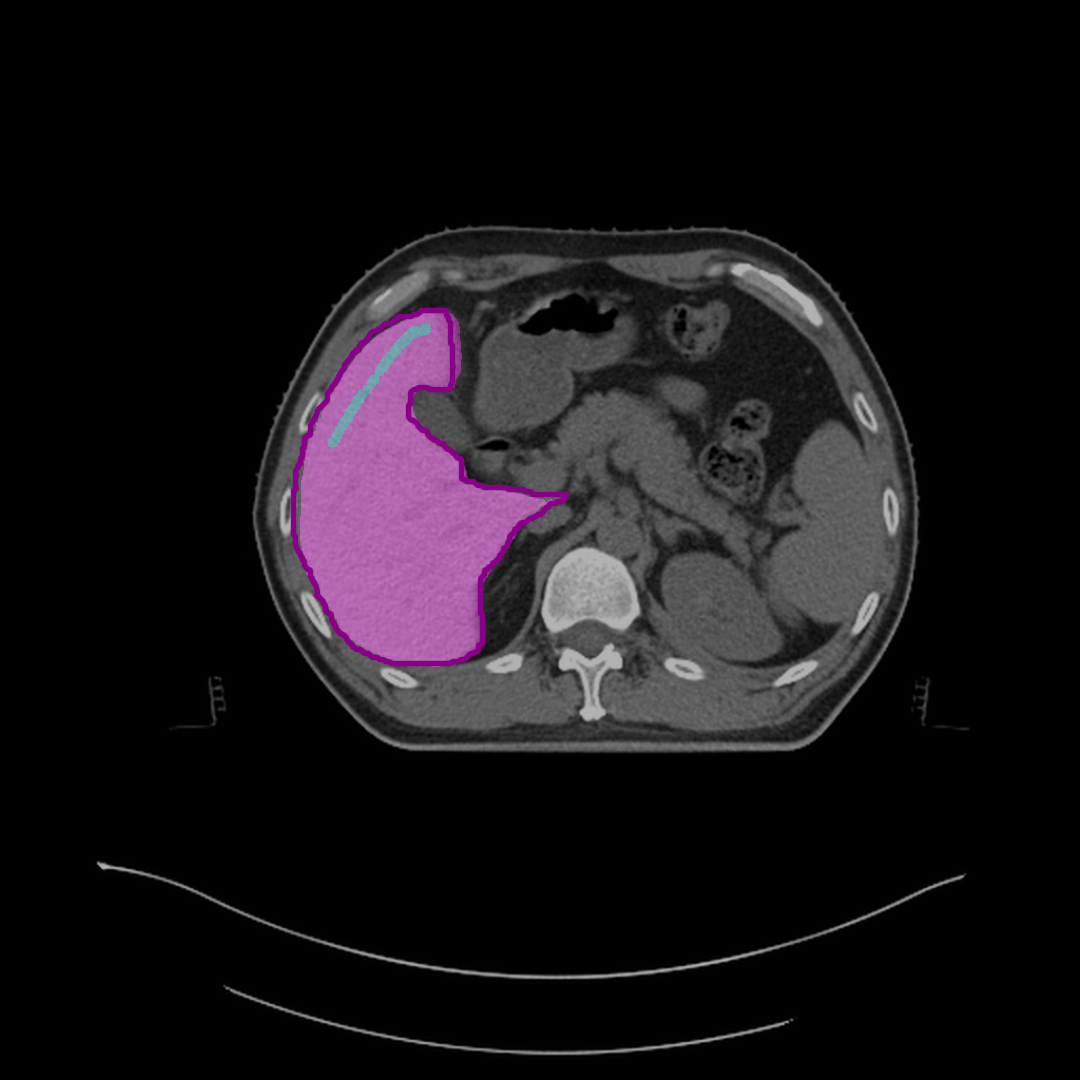}
  \caption{Mask after full correction.}
  \label{fig:short-i-ap}
  \end{subfigure}
  
  \caption{Steps to correct segmentation masks for various abdominal organs, such as the spleen, left kidney, and liver, on different CT slices. Each subfigure shows the outline of reference segmentation contours, the predicted segmentation mask, and gaze points (blue) used for predictions based on gaze.}
  \label{fig:vis-steps-9}
\end{figure*}





\end{document}